\documentclass[letterpaper, 10 pt, conference]{ieeeconf}  

\IEEEoverridecommandlockouts                              

\usepackage{amsmath} 

\usepackage[utf8]{inputenc} 
\usepackage[T1]{fontenc}    
\usepackage{hyperref}       
\usepackage{url}            
\usepackage{booktabs}       
\usepackage{amsfonts}       
\usepackage{nicefrac}       
\usepackage{microtype}      
\usepackage{xcolor}         
\usepackage{graphicx}
\usepackage{amsmath}
\usepackage{cleveref}
\usepackage{xfrac}
\usepackage{multirow}
\usepackage{dsfont}
\usepackage[style=ieee]{biblatex} 
\usepackage{adjustbox}
\usepackage{caption}

\usepackage{caption, subcaption, floatrow}
\usepackage{pifont}
\newcommand{\xmark}{\ding{55}}%

\newcommand{\bfit}[1]{\textbf{\emph{#1}}}

\newcommand\SF{\mathcal F}

\newcommand\SC{\mathcal{C}}
\newcommand\SL{\mathcal{L}}
\newcommand\SM{\mathcal{M}}

\newcommand\ST{\mathcal T}

\newcommand\ie{i.e.}
\renewcommand{\baselinestretch}{0.99} 

\DeclareMathOperator*{\argminA}{arg\,min} 

\title{\LARGE \bf Leveraging Cycle-Consistent Anchor Points for Self-Supervised RGB-D Registration}

\author{  \authorblockN{Siddharth Tourani\authorrefmark{1}\authorrefmark{4}, Jayaram Reddy\authorrefmark{2},  Sarvesh Thakur\authorrefmark{2},\\ K Madhava Krishna\authorrefmark{2}, Muhammad Haris Khan\authorrefmark{4}, N Dinesh Reddy\authorrefmark{3}}
  \authorblockA{\authorrefmark{1}
  Computer Vision and Learning Lab, 
    University of Heidelberg,\\
    \authorrefmark{4}
  MBZUAI,
  \authorrefmark{2}
  RRC, IIIT Hyderabad,
  \authorrefmark{3}
  Amazon}
}

\begin{document}

\maketitle
\thispagestyle{empty}
\pagestyle{empty}

\begin{abstract} 
With the rise in consumer depth cameras, a wealth of unlabeled RGB-D data has become available. This prompts the question of how to utilize this data for geometric reasoning of scenes. While many RGB-D registration methods rely on geometric and feature-based similarity, we take a different approach. We use cycle-consistent keypoints as salient points to enforce spatial coherence constraints during matching, improving correspondence accuracy. Additionally, we introduce a novel pose block that combines a GRU recurrent unit with transformation synchronization, blending historical and multi-view data. Our approach surpasses previous self-supervised registration methods on ScanNet and 3DMatch, even outperforming some older supervised methods. We also integrate our components into existing methods, showing their effectiveness. 
\end{abstract}


\section{INTRODUCTION}

RGB-D cameras are a rich source of information for scene understanding. They are especially useful for robotic tasks like: Simultaneous Localization and Mapping, drone navigation and object pose estimation. The growing use of such cameras has led to a substantial influx of RGB-D data lacking ground-truth pose information.

Typically, pose information for RGB-D data is subsequently derived using SfM pipelines~\cite{schonberger2016structure}, which can introduce noise and encounter optimization challenges, particularly in feature-scarce environments. The noisy pose information one may obtain  from such methods can lead to catastrophic failures in downstream robotic tasks.


 In this paper we investigate the use of salient portions of a scene to improve RGB-D registration. Existing self-supervised methods mostly leverage either feature similarity or geometric information available from the depth modality to perform this task. In our method, we exploit an under-exploited source of information, the salient portions of a scene. We assume the salient points of a scene are easily recognized in multiple views and leverage spatial relations with these points to constrain the difficult correspondence search process, yielding improved registration performance. We do so by incorporating a spatial coherence cost into the correspondence matching problem. 

\begin{figure}[h!]
    \centering
    \includegraphics[scale=0.10]{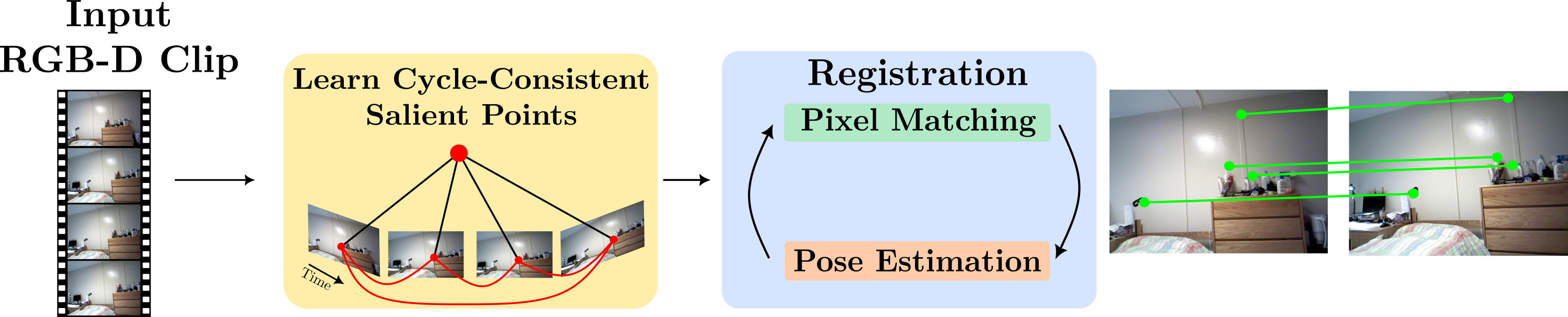}
    \caption{We propose to initially learn cycle-consistent salient points from RGB-D clips to substantially improve registration accuracy by subsequently constraining the correspondence estimation process.}
    \label{fig:teaser}
\end{figure} 

 Assuming these salient points are accurately localized, we exploit the fact that for correct correspondences the relative distance of a 3D point to these salient points should be transformation invariant. 
 
 Our proposed method is trained on RGB-D video clips without any ground-truth. An assumption in our method is that the salient points we recognize are cycle-consistent, \ie they are easily learned via a basic cycle-consistency loss and visible in every view. \textbf{We refer to these salient points as anchor points.}

Our method consists of an initial anchor point matching stage, in we learn anchor points for each video clip during training. We then incorporate these anchor points in the correspondence matching problem via a spatial coherence cost. As our method takes as input multiple RGB-D frames, for pose estimation we combine a GRU pose optimizer similar to the one proposed in~\cite{teed2021droid} with a pose synchronization module. The GRU unit incorporates past information via its hidden state, while pose synchronization leverages pose composition constraints across views yielding accurate pose estimates that are consistent across views.

Our contributions to the correspondence estimation and pose estimation modules lead to substantial improvements on ScanNet~\cite{dai2017scannet} and 3DMatch~\cite{zeng20173dmatch} datasets leading to new state-of-the-art for self-supervised RGB-D registration. 

To summarize our contributions are as follows:

\begin{itemize}
    \item We formulate a cycle-consistent keypoint matching module. The keypoints learned via this module impose spatial constraints on the correspondence estimation problem, improving registration.
    \item We propose a RANSAC-free approach for pose estimation, that combines historic information via a GRU unit with pose compositional consistency across views via transformation synchronization.
    \item We show through  experiments and ablation studies the benefit of our proposed modules. Our method achieves a new state-of-the-art among self-supervised RGB-D registration methods and approaches performance close to strong supervised baselines in terms of correspondence accuracy.
\end{itemize}

\begin{figure}[h]
    \centering
    \begin{subfigure}{0.43\textwidth}
        \centering
        \includegraphics[width=\textwidth]{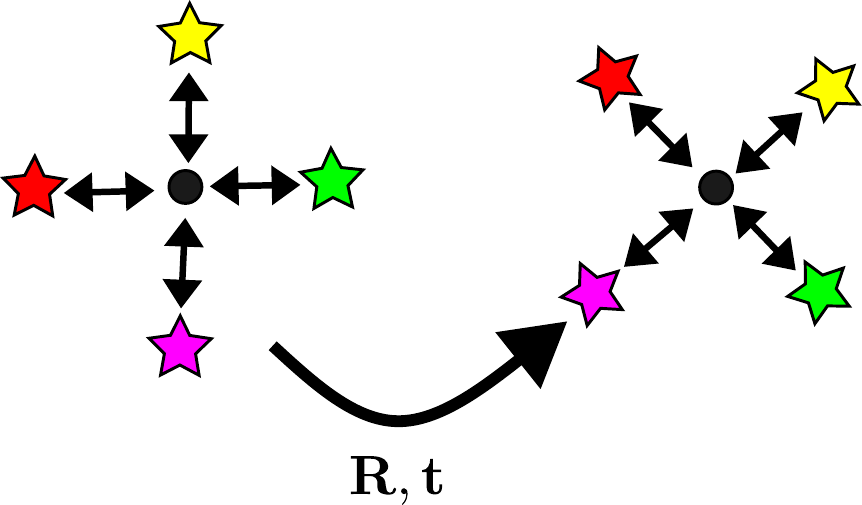}
        \caption{Spatially Coherent}
        \label{fig:subfig-spatail-coh}
    \end{subfigure}
    \hfill
    \begin{subfigure}{0.43\textwidth}
        \centering
        \includegraphics[width=\textwidth]{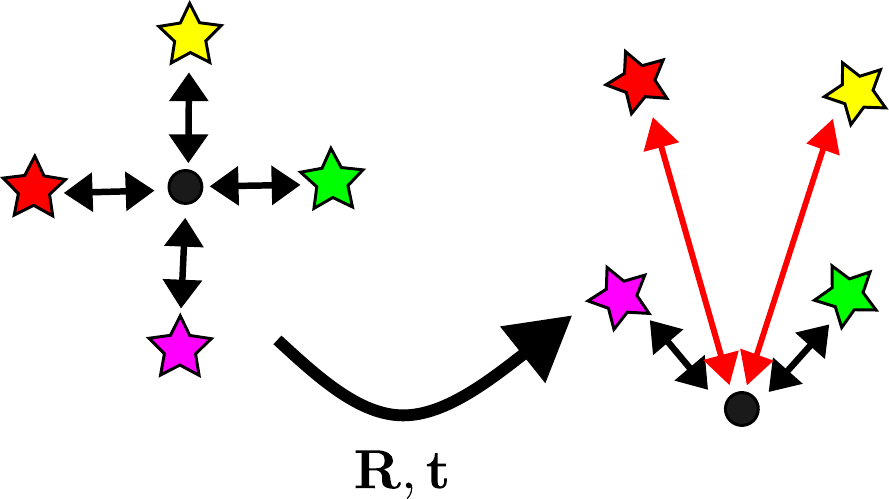}
        \caption{Spatially Incoherent}
        \label{fig:subfig2}
    \end{subfigure}
    \caption{ \textbf{Illustration of Spatial Coherence} The uniquely colored stars are anchor points, the black dot a correspondence and  arrows indicate distance. (a) shows a spatially coherent correspondence. In the transformed view, the distances between the correspondence and the anchor point are roughly identical. (b) shows a spatially incoherent correspondence, with the red arrows indicating distances that violate spatial coherence.}
    \label{fig:spatial-coherence}
\end{figure}
\section{Related Works}
\emph{We differentiate between point cloud registration and RGB-D registration} as they use different input modalities.
\paragraph{\textbf{Correspondence Estimation and Registration}} Classical point cloud registration techniques heavily relied on manually engineered features~\cite{Pomerleau2015ARO} or assumed perfect correspondence~\cite{Besl1992ICP}, but their effectiveness was bounded by the expressivity of these features. In upgrading point cloud registration methods to the deep learning era, learned counterparts have been proposed for the different components of the registration pipeline. These include learned keypoint descriptors~\cite{choy2019fully,Bai2020D3FeatJL,deng2018ppf}, correspondence estimation~\cite{choy2020deep,Yew2020RPMNetRP,yuan2020deepgmr,Qin2022GeometricTF,Wang2019DeepCP,Huang2020PREDATORRO}. Amongst the supervised RGB-D registration approaches proposed in the literature are~\cite{gojcic2020learning,gojcic2019perfect,Hertz2020PointGMMAN,le2019sdrsac,lu2019deepvcp,huang2020feature}.

While cycle-consistency has been used previously for correspondence estimation~\cite{jabri2020space,bian2022learning}, these methods impose cycle-consistency on all pixels in a video clip. We instead use cycle-consistency to localize salient points of a scene, which we then use as inputs to our correspondence estimation problem. \cite{bai2021pointdsc} uses within-frame spatial constraints  similar to us, but they do so to prune outliers as opposed to learn anchor points.

 \noindent\paragraph{\textbf{Self-Supervised Registration}} 
 In addition to the supervised methods mentioned above, there have been self-supervised and unsupervised RGB-D registration methods proposed~\cite{elbanani2021unsupervisedrr,elbanani2021byoc,elbanani2023syncmatch}. These serve as direct comparisons to our RGB-D registration methods.  All of these methods estimate correspondences using a weighted version of Lowe's ratio test~\cite{lowe2004distinctive} for correspondence estimation and the Kabsch algorithm~\cite{Kabsch1976ASF} for relative pose estimation between point cloud pairs. These methods are typically supervised by a single weighted \emph{L2-registration loss}. 
In a different category, self-supervised point cloud registration methods,  have been proposed as well~\cite{shen2022reliable,Mei2023UnsupervisedDP,Huang2022UnsupervisedPC,mei2023unsupervised,kadam2020unsupervised}.

 \begin{figure*}[t!]
    \centering
    \includegraphics[scale=0.20]{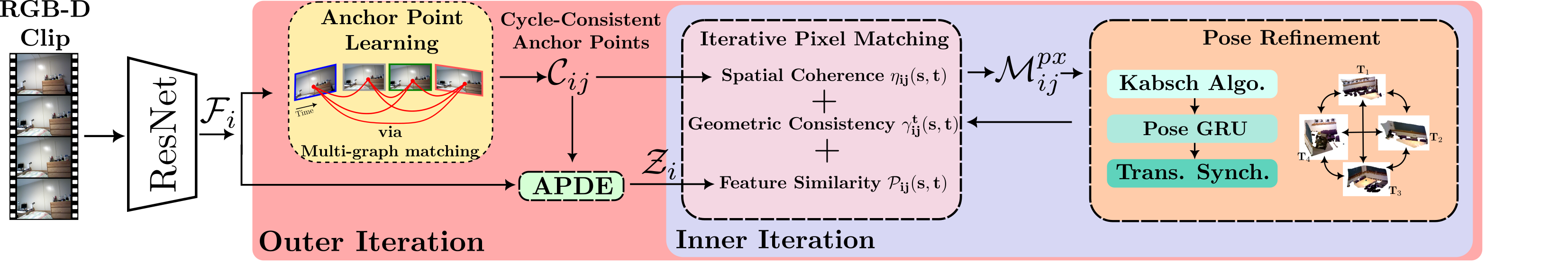}
    \caption{ \textbf{Overview of our Method} Features are extracted from an RGB-D clip a via ResNet backbone. They are then used to learn a set of anchor points that are cycle-consistent across frames from the clip. These cycle-consistent anchor points are input into the pixel matching module. The pixel matching module and pose refinement module iterate feeding into each other. This is termed the inner iteration. The outer iteration happens periodically to update the anchor point locations. ADPE stands for Anchor Point Distance Embedding.}
    \label{fig:method}
    \vspace{-1mm}
\end{figure*}
\section{Method} 
\label{sec:method}

From each video clip, we learn a set of cycle-consistent anchor points $\SC_{ij}$ by solving a multi-graph matching problem, where i, j refers to the frame indices. These points are used to enforce spatial constraints to pixel matching problem, which yields a set of soft-correspondences $\SM_{ij}^{px}$. The soft correspondences  are in-turn used for pose-estimation and refinement. The estimated pose information is fed into pixel matching block as geometric consistency cost. Pixel matching and pose information are iterated over. This we define  as the inner iteration. As the estimated anchor points can be noisy, we periodically update the anchor points as well. This we define as the outer iteration. Our method is outlined in \Cref{fig:method} and we explain each module in detail below.

\textbf{Notation} $\mathbb{I}$ is the identity matrix.  $i,j$ throughout the paper are indices of the frames, while $r, s, k, l$ refer to the matched patches or pixels and $t$ denotes time.

\subsection{Feature Extraction} We use a pre-trained ResNet-18~\cite{he2016deep} to extract local coarse (at $1/4$ resolution) and fine-level features (at $1/2$ resolution), we downsample the fine-level features via average pooling to the coarse level resolution and concatenate them along the channel dimension yielding features $\SF_i$ that combine both high and low-frequency information.

\subsection{Anchor Point Learning / Matching}
\label{sec:correspondence} 
We utilize the memory efficient matching strategy used in LofTr~\cite{sun2021loftr} and flatten the features $\SF_i$ to 1-D vectors. We construct matching problem by computing the dot product of feature maps between frames,  \ie the score matrix of  frames $i,j$ is:
\begin{equation}
    \mathcal{S}_{ij} = - \left\langle \SF_{i}, \SF_{j} \right\rangle
    \label{eqn:score-matrix-coarse}
\end{equation}
We do this for all pairs of frames. These score matrices are then optimized via \emph{Sinkhorn normalization}~\cite{sinkhorn1967concerning} yielding soft matchings $\SM_{ij}$ between pairs of frames. To allow for partial matching arising from different field of view, occlusions and missing depth we add slack row and column vectors to allow for non-assignment, as in~\cite{sarlin2020superglue}.

To localize the keypoints in higher resolution, we up-sample the feature map and correspondences. We convert these soft pairwise matches $\SM_{ij}$, into cycle consistent matches using the matrix factorization method proposed in~\cite{bernard2019synchronisation}. We use these cycle-consistent matches as anchor points in our pixel matching module, to impose addition spatial constraints on the correspondence estimation process. They are denoted by $\SC_{ij}$ for correspondences between frames $i$, $j$. The anchor point locations are stored as parameters and periodically optimized by minimizing their reprojection error.

\subsection{Pixel Level Matching}

\noindent\textbf{Anchor Point Distance Encoding}
Relative distance encoding has been shown to encode $\mathbb{SE}(3)$ invariant information within a point cloud and effective in supervised registration~\cite{Qin2022GeometricTF}. Incorporating the same encoding into our method, we found that unlike in~\cite{Qin2022GeometricTF} it gave only marginal benefit. This difference is probably because the self-attention module in~\cite{Qin2022GeometricTF} benefits from supervision and access to ground-truth for determining relevant regions within a point cloud. Conversely, our self-supervised approach operates on pseudo ground-truth.


Hence, our proposal is to capture the $\mathbb{SE}(3)$ invariant information by capitalizing on the stability of our anchor points and solely measuring distances relative to these anchor points. We thus define an \emph{anchor point distance embedding} $\mathbf{r_p}$ which we integrate into a self-attention module. 

Given a feature matrix $\SF_i \in \mathbb{R}^{N \times d}$ as input, the modified attention module outputs a feature matrix $\mathcal{Z}_i \in \mathbb{R}^{N \times d}$ which is the weighted sum of all projected features
\begin{equation}
\mathbf{z_p} = \sum_{q=1}^N a_{pq}(\mathbf{x_pW^V})
\end{equation}
where $a_{pq}$ is the normalized attention score computed by a row-wise softmax of $e_{pq}$.
\begin{equation}
    {e_{pq}} =\frac{(\mathbf{x_pW^Q+}{\color{red} \mathbf{r_{p}}\mathbf{W^R}})(\mathbf{x_q W^K + } {\color{red} \mathbf{r_{q}W^R})}}{\sqrt{d}}
\end{equation}
Here $d$ is the feature dimension. $\mathbf{W^Q}$, $\mathbf{W^K}$, $\mathbf{W^V}$ and $\mathbf{W^R}$ are projection matrices. The modified elements of the self-attention block are shown in {\color{red} red}. We have modified the self attention equation based on the best practices espoused in~\cite{wu2021rethinking}, which explores the different ways to do this.

Let the average distance between $\mathbf{p}$ and the anchor points in its frame  be$\rho_{pq}$. $\mathbf{r_p}$ is computed by applying a sinusoidal function~\cite{vaswani2017attention} on $\sfrac{\rho_{pq}}{\sigma_d}$, where $\sigma_d$ is a hyper-parameter used to tune the sensitivity to distance variations.
We upsampled the feature maps $\SF_i$ to $1/2$ half resolution and pass them through the anchor point aware self attention block and get as output feature maps $\mathcal{Z}_i$. These are then input into the pixel matching module.

\textbf{Spatial Coherence Cost} We define spatial consistency as the consistency within frame of a point relative to the other points in the same frame. Intuitively speaking, it encodes transformation invariant information within frame  which  can be used as a cost to penalize spatially inconsistent correspondences across frames. Incorporating some form of spatial consistency has been shown to improve point cloud registration in~\cite{bai2021pointdsc,zhang2022pcr}. 

We instead compute a spatial consistency cost function by measuring distance to anchor points, as follows:
\begin{eqnarray}
d_{rs} =  \sum_{(k,l) \in \SC_{i,j}} \mathtt{abs(} \| \mathbf{x}_r - \mathbf{x}_k \| - \| \mathbf{x}_s - \mathbf{x}_l \| \mathtt{)}\\
\eta_{ij}(r,s) =-\mathtt{exp}(\frac{-d_{rs}^2}{\sigma_{rs}^2}) \label{eqn:sc}
\label{eqn:rel-dist}
\end{eqnarray}
Here, $(k,l)$ are the anchor points in $\SC_{ij}$. \texttt{abs} is the absolute value. $\mathbf{x}_r$ and $\mathbf{x}_s$ are the 3D point corresponding to $r$, $s$ and $\sigma_{rs}$ is a learned hyper-parameter.

$\eta_{ij}(r,s)$  tends to 1 when the sum  of distances between $\mathbf{x}_r$ and other matched points in frame $i$ is close to sum  of distances between $\mathbf{x}_s$ and other matched points in frame $j$ and 0 when the difference is larger. \Cref{fig:spatial-coherence} gives an intuitive illustration of the spatial coherence cost. 

\textbf{Iterative Pixel-Level Matching}
The estimated anchor points  allow us to incorporate spatial coherence costs and geometric costs into the matching process, providing additional sources of information to complement feature similarity. The pixel level matching block is used to get more  and better localized correspondences . 

To keep the pixel level matching tractable, we restrict the matching to $w\times w$ windows centered around each anchor point. These are also regions of higher confidence as they center on anchor points.

Let $\mathcal{Z}_i(r)$, $\mathcal{Z}_i(s)$ be the feature maps of the $w\times w$ windows at points $r$ and $s$ in frames $i$ and $j$ respectively. We initialize the cost matrix between matched points $(r,s)$ of frames $(i,j)$ as follows:
\begin{equation}
    \mathcal{P}_{ij}(r,s) = - \left\langle \mathcal{Z}_i(r), \mathcal{Z}_j(s) \right\rangle
    \label{eqn:fine-score-matrix-initial}
\end{equation}
We iterate between the pixel-level matching block and pose update block matching as their outputs feed into each other leading to convergence to a fixed point, which hopefully fits the underlying data. 

The pixel-level matching problem at inner iteration $t$ is defined as 
\begin{equation}
     \mathcal{D}^t_{ij}(r,s) = - \mathcal{P}_{ij}(r,s)  - \eta_{ij}(r,s) - \gamma_{ij}^t(r,s)\label{eqn:iter-px-match}
\end{equation}
The geometric cost function, $\gamma^t_{ij}(r,s)$ is the Sampson error normalized to unit norm. It takes as input the relative transformations between frames $i$ and $j$. For exact correspondences and relative pose the value is zero and increase with error. These are initially computed from the anchor points via~\cite{Kabsch1976ASF} and subsequently refined by the pose estimation block.

We optimize the score matrices $\mathcal{D}^t_{ij}(r,s)$ via \emph{Sinkhorn normalization}. $\SM^{px}_{ij}$ denotes the soft correspondences \ie for frames $(i,j)$, output by the Sinkhorn algorithm. 

\subsection{Iterative Pose Refinement} 
\label{sec:pose-est} 
We leverage both the historical information via a GRU unit and transformation consistency across views via transformation synchronization to improve pose estimation accuracy.

Given the soft correspondences $\SM_{ij}^{px}$,  transformation update $\delta\mathcal{T}\in \mathbb{SE}(3)$ is obtained by minimizing the following weighted mean-squared error:
\begin{equation}
\small
    \delta\mathcal{T}_{ij}^* = \argminA_{\delta\mathcal{T}} \sum_{(r,s) \in \SM^{px}_{ij}} w_{rs} \| \mathbf{x}_{r} - \mathcal{T}_{ij}(\mathbf{x}_s) \|; \label{eqn:kabsch}
\end{equation}
This is done via a modified Kabsch~\cite{Kabsch1976ASF} algorithm, commonly used in differentiable point cloud registration. $w_{rs} = \text{softmax}(\mathcal{D}^t_{ij}(r,s))$ weighs the L2-objective.

Inspired by~\cite{teed2021droid}, we add a GRU based pose update block to incorporate information from past time steps into the pose estimation process. Our GRU block predicts relative pose updates $\Delta \ST_{ij}$ as 6D vectors (using the representation in~\cite{zhou2019continuity}). We update the transformations via an $\mathbb{SE}(3)$ retraction.
\begin{equation}
\mathcal{T}_{ij}^{t+1} = \mathcal{T}^{t}_{ij} \cdot \text{Exp}(\Delta \ST_{ij})
\end{equation}
Here Exp(.) is the $\mathbb{SE}(3)$ retraction.

After each update we additionally perform a single transformation synchronization iteration to average out the pose errors across frames. We use the power iteration algorithm used in~\cite{elbanani2023syncmatch} as its numerically stable. The transformation terms are weighed by the mean of the soft correspondences for each of the frame pairs. 


\textbf{Inner Iteration} The inner iteration is used to iterate between the pixel-level matching and the pose update block to refine and increase the number of correspondences. Each inner iteration is performed 20 times.

\textbf{Outer Iteration} The outer iteration iterates over the entire pipeline, \ie the anchor point learning, pixel matching and pose refinement. It is done  3 times per batch. The outer iteration lets for updating the anchor points and their locations. We do this by minimizing the reprojection error via~\cite{pineda2022theseus}. Optimizing anchor point locations in-turn improves downstream task accuracy.

Please refer to appendix for addition details.



\subsection{Supervision}
We use the registration loss which minimizes the weighted residual error of the estimated correspondences using the estimated alignment. Given a pair of frames $(i,j)$, we compute it as follows:
\begin{eqnarray}
    \SL_{reg}^{3D}(i,j) = \sum_{(r,s) \in \SM^{px}_{ij}}w_{rs}  \|\mathbf{x}_s - \mathcal{T}_{ij}(\mathbf{x}_r) \|_2^2 \label{eqn:pose-estimation}
\end{eqnarray}
 $w_{rs}=\text{softmax}(\mathcal{D}^t_{ij}(r,s))$ is a weighing term. 

\noindent\textbf{Cycle Consistency Loss} We encourage cycle-consistency by combining orthogonality ($\SL_{orth}$) and  bijectivity ($\SL_{bij}$)
\begin{eqnarray}
    \SL_{orth}(i,j) = \| \mathcal{S}_{ij}\mathcal{S}^{T}_{ij} - \mathbb{I} \|_F + \| \mathcal{S}_{ij}^{T}\mathcal{S}_{ij} - \mathbb{I} \|_F \\
    \SL_{bij}(i,j) = \| \mathcal{S}_{ij}\mathcal{S}_{ji} - \mathbb{I} \|_F + \| \mathcal{S}_{ji}\mathcal{S} - \mathbb{I} \|_F 
\end{eqnarray}
This formulation of cycle-consistency is simpler than the formulation used in~\cite{jabri2020space} and is applied for anchor point learning only.

\subsection{Test Time}
At test time, we do not store or use anchor points. Our correspondence estimation module still consists of the two stages. At the initial anchor point learning stage, the matching problem is the same as at train time (see~\cref{eqn:score-matrix-coarse}). Inspired by~\cite{wang2020learning}, for every anchor point match $(r,s)$ between frames $(i,j)$, we locate its position $(\hat{r},\hat{s})$ in the fine scale feature maps. We then crop two $w \times w$ local windows of the feature maps $\mathcal{Z}_i(\hat{r})$, $\mathcal{Z}_j(\hat{s})$  centered on  $(\hat{r},\hat{s})$. We correlate the center feature of $\mathcal{Z}_i(\hat{r})$ with all the features of $\mathcal{Z}_j(\hat{s})$ and normalize, giving us a probability of matching each pixel in the neighborhood of $\hat{s}$ to the pixel $\hat{r}$. We take the expectation over this heat-map to get the final correspondence. 
For pose estimation, we use only the Kabsch algorithm~(\cref{eqn:kabsch} and discard the GRU and transformation synchronization.

\section{Results}
We follow the evaluation protocol proposed in UR$\&$R~\cite{elbanani2021unsupervisedrr}. 

\paragraph{Datasets} We evaluate on 3DMatch~\cite{zeng20173dmatch} and the ScanNet~\cite{dai2017scannet} \emph{v2} split.  ScanNet provides RGB-D videos of 1513 scenes. 3DMatch provides 72 train sequences.
Details of the datasets can be found on their websites and in the appendix.


\begin{table}[h]
\setlength{\tabcolsep}{2pt}
\scriptsize
    \centering
    \begin{tabular}{ccccccc|cccc}
        \multirow{2}{*}{Method} & & &  \multicolumn{4}{c}{Angular Error} & \multicolumn{4}{c}{Translation Error}  \\
         & Train   & Sup.  & \multicolumn{2}{c}{\underline{\text{Accuracy $\uparrow$}}} & \multicolumn{2}{c}{\underline{\text{Error $\downarrow$}}} & \multicolumn{2}{c}{\underline{\text{Accuracy $\uparrow$}}} & \multicolumn{2}{c}{\underline{\text{Error $\downarrow$}}}  \\
         & Set &  & $5^\circ$ & $10^\circ$ & Mean & Med. & $5$ & $10$ & Mean & Med. \\
         \hline
         \\
   \texttt{3DMVR} & \parbox[t]{2mm}{\multirow{8}{*}{\rotatebox[origin=c]{90}{Trained on 3DM}}}
 & \checkmark & 81.1  &  89.3  & 9.4  &  1.8  &  54.5  &  76.2 &  18.4  &  4.5 \\
   \texttt{DGR} & & \checkmark &  87.7  & 93.2  & 6.0  & 1.2  & 69.0 & 83.1 &  11.7 &  2.9 \\
    \texttt{Geom.Tr.} & & \checkmark &  \bfit{98.3} & \bfit{99.6} & \bfit{1.1} & \bfit{0.6} & \bfit{91.8} & \bfit{96.5} & \bfit{2.4} & \bfit{0.8}\\
    \\
\texttt{UR\&R} &  & \xmark & 87.6  & 93.1 &  4.3 &  1.0 &  69.2 &  84.0 &  9.5 &  2.8 \\    
\texttt{BYOC}~\cite{elbanani2021unsupervisedrr} &  & \xmark & 66.5 &  85.2 & 7.4 &  3.3 &  30.7 &  57.6 & 16.0 & 8.2\\
     \texttt{LLT}~\cite{wang2022improving} &  & \xmark & 93.4 & 96.5 & 3.0 & \underline{0.9} & 76.9 & 90.2 & 6.4 & 2.4\\
\texttt{SyncM}~\cite{elbanani2023syncmatch} &  & \xmark & 93.4 & 97.6 & 2.8 & \textbf{0.7}  & 76.6 & 89.9 & 7.1 & 2.6 \\
    \texttt{Ours} &  & \xmark &  \textbf{95.6} &  \textbf{98.1} &   \textbf{2.4} & \textbf{0.7} & \textbf{81.5} & \textbf{92.3} & \textbf{3.7} & \textbf{1.9}\\
\\
    \texttt{UR\&R}~\cite{elbanani2021unsupervisedrr} & \parbox[t]{2mm}{\multirow{5}{*}{\rotatebox[origin=c]{90}{Trained on SN}}} & \xmark &  92.7 & 95.8 & 3.4 & 0.8 & 77.2 & 89.6 & 7.3 & 2.3\\
    \texttt{BYOC}~\cite{elbanani2021byoc} & & \xmark & 86.5 & 95.2 & 3.8 & 1.7 & 56.4 & 80.6  & 8.7 & 4.3\\
   \texttt{LLT}~\cite{wang2022improving} & & \xmark & \underline{95.5} & \underline{97.6} &   2.5 & 0.8 & 80.4  & 92.2  & 5.5 & 2.2\\
    \texttt{SyncM}~\cite{elbanani2023syncmatch}& & \xmark & 95.4 & 97.5 & \underline{2.4} & \underline{0.7} & \underline{81.3} & \textbf{93.8} & \underline{5.4} & \underline{1.9}\\
    \texttt{Ours} & & \xmark & \textbf{97.1} & \textbf{98.2} & \textbf{1.9} & \textbf{0.6} & \textbf{85.9} & \underline{93.6}  & \textbf{3.9} & \textbf{1.8}\\
    \end{tabular}
\caption{\textbf{Registration Results On ScanNet}  Best results are bold and italicized for supervised methods. Best results and next best are bold and underlined for un/self-supervised methods. The train set can be SN (ScanNet) or 3DM (3D Match). “Sup.” indicates whether the method is supervised or not.  }
    \label{tab:reg}
\end{table}
\paragraph{Training Details} Following~\cite{elbanani2021unsupervisedrr}, we generate sequences consisting of 6 images sampled 20 frames apart as in~\cite{elbanani2021unsupervisedrr}. Images are reshaped to $256 \times 256$px in size. Our model is optimized with the Adam~\cite{kingma2014adam} optimizer using a learning rate of $5 \times 10^{-4}$ and momentum parameters of  $(0.9, 0.99)$. Hard matches are obtained by thresholding $\SM^{px}_{ij}$ at test time. A few iterations of anchor point learning are done to learn some anchor points before moving to pixel matching and pose estimation.




\subsection{Adding Spatial Coherence To Other Methods}
\label{subsec:add-sc}
To quantify the impact of spatial coherence,  we integrate it into other methods. A non-trivial task. Instead of learning anchor points, we sample correspondences present in all input frames using ground-truth poses. These approximately 50 anchor points, similar in number to those learned by our pipeline, helps us introduce spatial coherence costs into other methods. Due to their ground-truth origin, these anchor points offer higher accuracy than those generated by our self-supervised pipeline, potentially providing greater benefits.

\subsection{Registration Accuracy on ScanNet} We first evaluate our approach on RGB-D registration accuracy. The transformation is represented by a rotation matrix \textbf{R} and translation vector \textbf{t}. We use the commonly used angular error and translation error (see appendix or~\cite{elbanani2021unsupervisedrr} for formulas) as evaluation metrics for registration accuracy.

\noindent\textbf{Baselines} We compare against the following  un/self-supervised methods: \texttt{BYOC}~\cite{elbanani2021byoc}\footnote{we modify the backbone from ResNet-5 to ResNet-18 to make the visual backbone similar to the other methods.} , \texttt{UR\&R}~\cite{elbanani2021unsupervisedrr}, \texttt{LLT}~\cite{wang2022improving} and \texttt{SyncMatch}~\cite{elbanani2023syncmatch}. We also show comparisons against older supervised methods \texttt{3DMVR}~\cite{gojcic2020learning} and \texttt{DGR}~\cite{choy2020deep} our method outperforms. Additionally, to put into context our method compared to the supervised state-of-the-art, we show results on the current state-of-the-art \texttt{Geom.Tr.}~\cite{Qin2022GeometricTF}.

Table~\ref{tab:reg} shows results of registration on ScanNet and 3DMatch.  \texttt{Ours} outperforms other un/self-supervised methods on both 3DMatch and ScanNet by wide margins in most metrics. We analyze them in detail in the ablation study.

\begin{table}[h]
\scriptsize
    \centering
    \begin{tabular}{cccccccc}
        Method & Inputs  &  \multicolumn{3}{c}{3D Corres.} & \multicolumn{3}{c}{2D Corres.}   \\
       &  & 1cm & 5cm & 10cm & 1px & 2px & 5px\\
       \hline
         \multicolumn{8}{c}{ \textbf{Supervised features with trained matching}}\\
         \texttt{SG} & I & 8.7 & 62.4 & 78.7 & 2.5 & 9.0 & 36.9\\
         \texttt{SG+SC} & I+D & 16.8 & 78.8 & 87.7 & 5.9 & 19.7 & 58.9\\
           \texttt{LoFTr} & I & 16.0 & 72.2 & 84.6 & 5.6 & 18.5 & 55.5\\
         \texttt{LoFTr+SC} & I+D  & 24.5 & \underline{82.1} & \underline{92.1} & \textbf{9.5} & \textbf{27.2} & \textbf{63.1}\\
         \\
         \multicolumn{8}{c}{\textbf{Unsupervised features with heuristic matching}}\\
         \texttt{BYOC} & D & 13.1 &  55.1  &  65.4 &  4.6 & 15.3 & 43.9 \\
         \texttt{UR\&R}  & I+D & 24.3 & 4.5 &  82.6 & 6.9 & 19.5 &  53.3  \\
         \texttt{LLT}  & I+D & 26.2 & 75.9 &  82.1 & 7.2 & 22.4 &  58.7  \\
         \texttt{SyncM} & I+D &  13.1 & 55.1 & 65.4 & 4.6 & 15.3 & 43.9  \\
         \texttt{SyncM+GART}  & I+D & \underline{26.8} & 76.5 &  84.4 & 7.5 & 23.5&  59.7  \\
         \\
         \multicolumn{8}{c}{\textbf{Unsupervised features with trained matching}}\\
        \texttt{Ours} & I+D & \textbf{31.2} & \textbf{84.3} & \textbf{92.7} & \underline{9.1} & \underline{25.3} & \underline{61.2} \\
    \end{tabular}
    \caption{ \textbf{Correspondence Inlier \% on ScanNet} For the inputs, I denotes image and D denotes depth.  \texttt{LofTr} and \texttt{SG} are originally image based correspondence methods, so incorporating depth information involves some assumptions. Also, adding \texttt{SC} to various methods improves correspondence estimation, quantifying its efficacy. \texttt{SyncM+GART} is a SyncM variant that incorporates geometric constraints.}
    \label{tab:fmr}
\vspace{-1mm}
\end{table}

\subsection{Correspondence Estimation} 
\noindent\textbf{Evaluation Metrics} We evaluate the estimated correspondences based on their 2D and 3D errors. We project the estimated correspondences into 3D for valid keypoints with depth using known depth and intrinsics. The ground truth transformations are used to align the keypoints and compute the 3D error and the 2D reprojection error. We extract 500 correspondences for all methods to allow for a meaningful comparison between precision values. \footnote{If a method produces $<500$ correspondences we use all of them.}

\noindent\textbf{Baselines} We compare against supervised and self-supervised  counterparts: SuperGlue~\cite{sarlin2020superglue} \texttt{SG}
attention-based matching algorithm built on top of SuperPoint, \texttt{LoFTr}~\cite{sun2021loftr} image based feature matching method, \texttt{UR\&R}, \texttt{BYOC}, \texttt{SyncM} and \texttt{LLT}.

To isolate the impact of our spatial coherence term, we also incorporate it  into the matching problems of \texttt{SG} and \texttt{LoFTr} to assess its effectiveness. This has the added effect of incorporating depth information into \texttt{SG} and \texttt{LofTr} which are RGB image feature matching methods. These modified methods are represented by \texttt{XXX + SC} where \texttt{XXX} is the method name. Refer to~\cref{subsec:add-sc} to better understand how these methods make use of anchor points. For \texttt{LofTr}, we use Sinkhorn normalization instead of the Dual SoftMax to easily  incorporate spatial coherence. We add the spatial coherence costs to the fine-matching stage similar to our method.
\Cref{tab:fmr} shows the results of correspondence estimation. While our method is still worse than strong supervised baselines in 2D correspondence, it is still relatively close to them and can be used as a suitable alternative to commonly used self-supervised correspondence estimation algorithms. Additionally, the effect of the spatial coherence cost (SC.) is strongly positive improving the percentage of correspondence inliers. Also note while these methods, assume different modalities (RGB or depth only). We do our best to modify them to use both when adding \texttt{SC}.

\subsection{Importance of individual modules}

\subsubsection{Ablation Study}
We ablate the following components of our pipeline: anchor point depth encoding (\texttt{DE}), spatial coherence costs (\texttt{SC}), geometric costs (\texttt{GC}), and the pose estimation block (\texttt{PE}). We measure the registration error and correspondence accuracy in 2D  (\emph{in pixels}) and 3D (\emph{in cms}). Results are averaged over all test sequences in ScanNet.

\begin{figure}[h]
     \centering
     \begin{subfigure}[b]{0.45\textwidth}
         \centering
         \includegraphics[width=\textwidth]{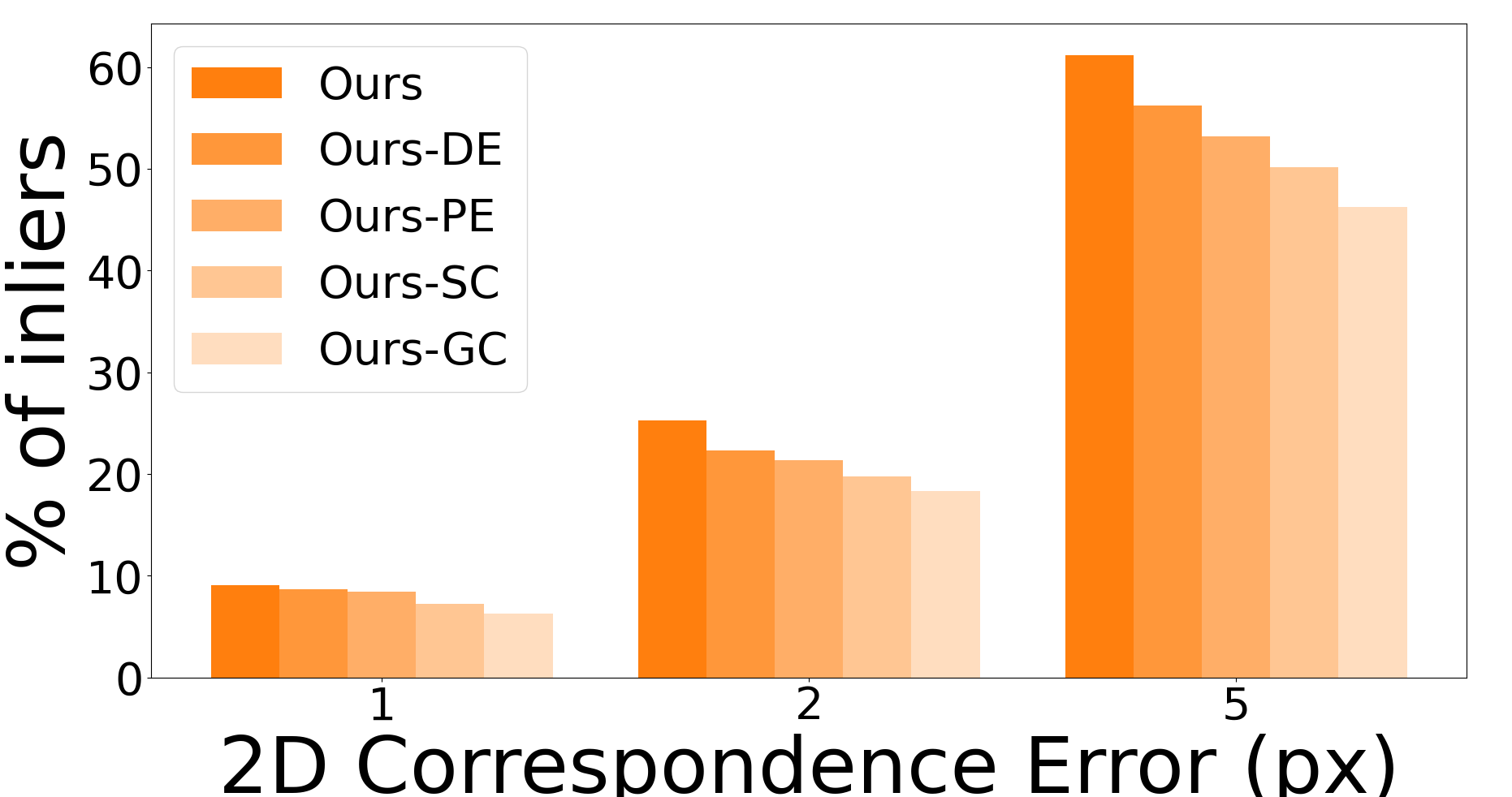}
         \caption{\small 2D Error (in px) }
         \label{fig:corr_2d}
     \end{subfigure}
     \begin{subfigure}[b]{0.45\textwidth}
         \centering
         \includegraphics[width=\textwidth]{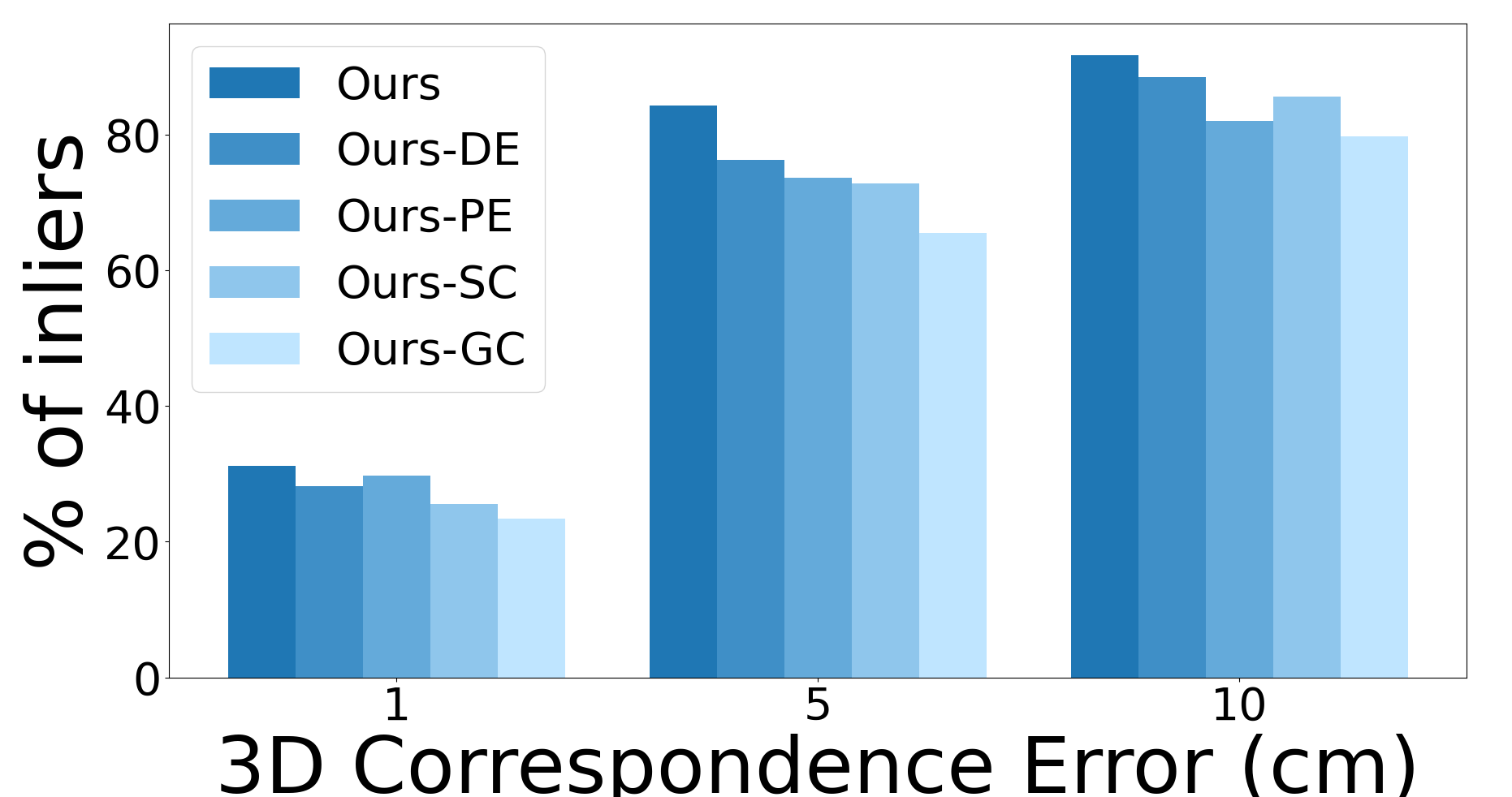}
         \caption{3D Error (in cm)}
         \label{fig:corr_3d}
     \end{subfigure}
\caption{ \textbf{Correspondence Ablation} Ablation Study for Inlier \% under various thresholds in 2D and 3D. While swapping or removing all four modules brings a decrease in inlier count, \texttt{GC} has the most significant effect followed by \texttt{SC}.}
\label{fig:corr_ablation}
\end{figure}

\Cref{fig:corr_ablation} and \cref{tab:rot_ablation}  shows the results of removing \texttt{DE}, \texttt{SC} \texttt{GC} and \texttt{PE} on correspondence error and rotation error respectively. Unsurprisingly \texttt{GC} which incorporates pixel specific information contributes the most to our method in correspondence estimation and registration accuracy.  However,  closely following it is \texttt{SC}, demonstrating our spatial coherence costs that make use of anchor points are also quite effective.

\begin{table*}[t]
\begin{tabular}{cccc}
\texttt{LLT} & \texttt{SyncM} &  \texttt{Ours} & \texttt{GT}\\
\hline
\includegraphics[width=0.17\textwidth]{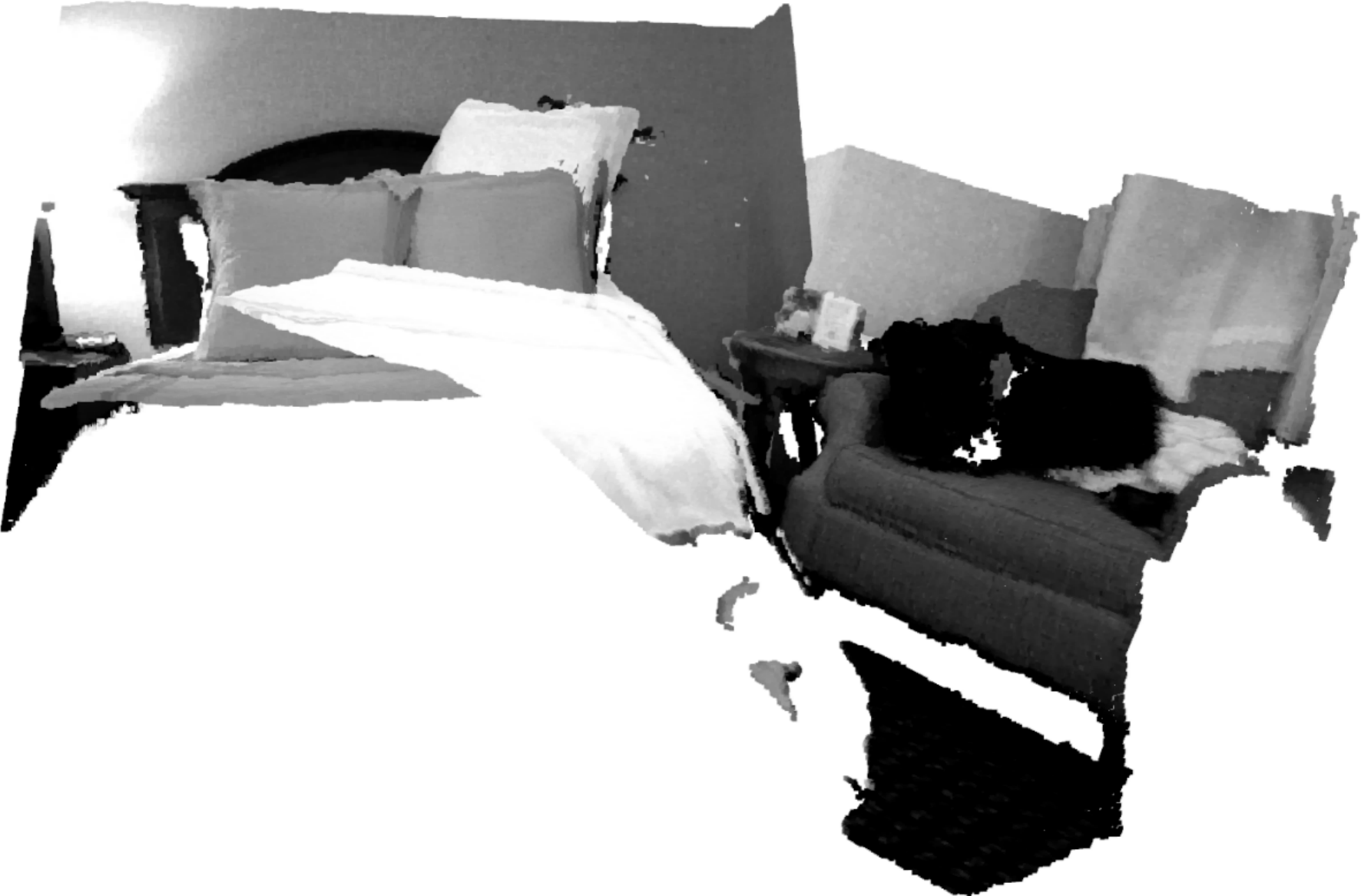} & \includegraphics[width=0.17\textwidth]{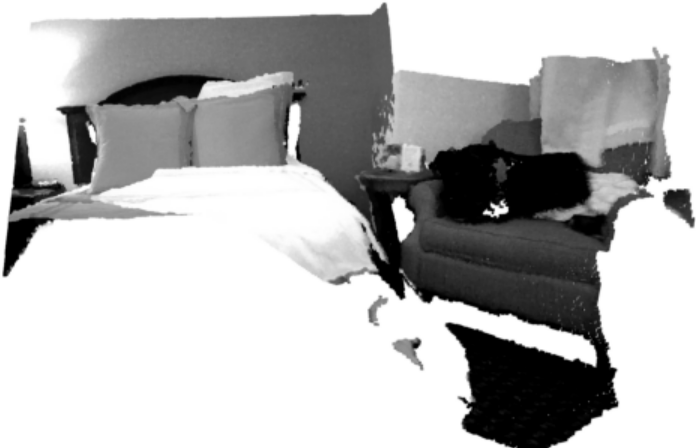} & \includegraphics[width=0.16\textwidth]{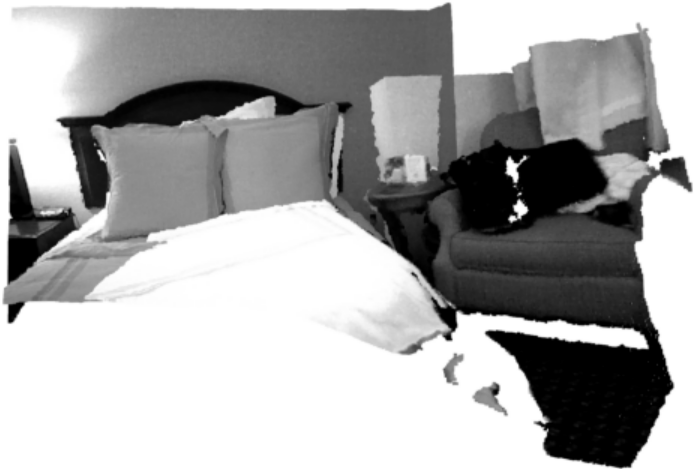} &\includegraphics[width=0.15\textwidth]{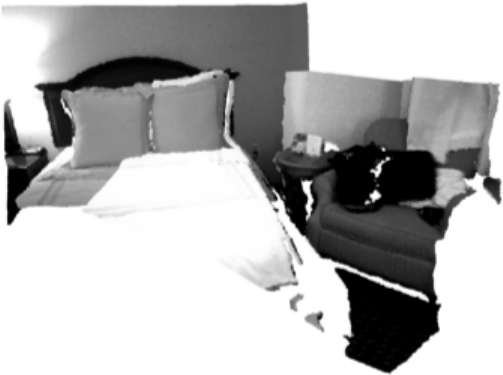} \\
$R_{er}=0.24$, $t_{er}=0.05$ & $R_{er}=0.1$, $t_{er}=0.12$ & $R_{er}=0.04$, $t_{er}=0.02$ & $R_{er}=0$, $t_{er}=0$\\
\\
\includegraphics[width=0.17\textwidth]{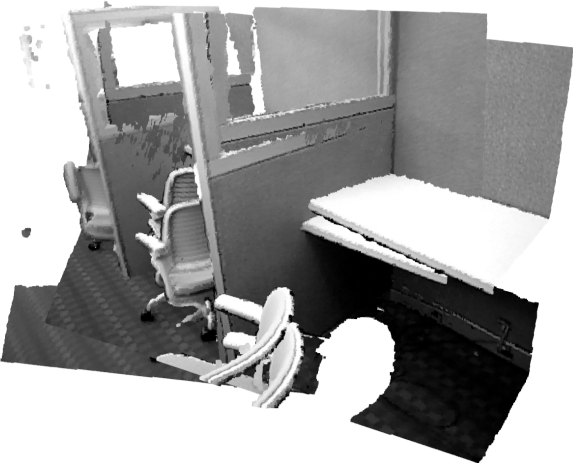} & \includegraphics[width=0.17\textwidth]{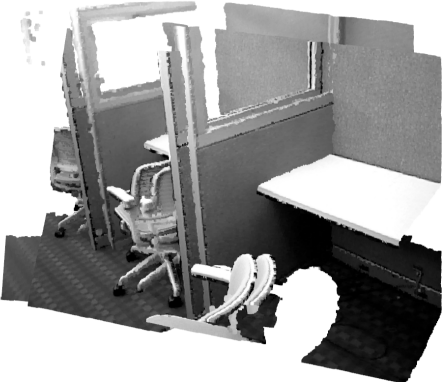} & \includegraphics[width=0.17\textwidth]{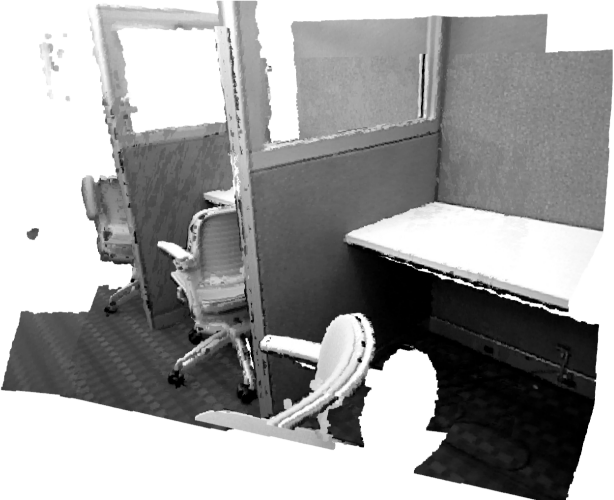} &\includegraphics[width=0.17\textwidth]{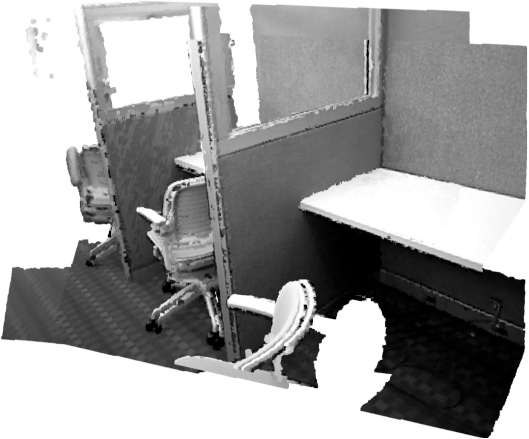} \\
 $R_{er}=0.09$, $t_{er}=0.21$ & $R_{er}=0.08$, $t_{er}=0.12$ & $R_{er}=0.05$, $t_{er}=0.08$ & $R_{er}=0$, $t_{er}=0$\\
\end{tabular}
\caption{\textbf{Qualitative Results on 3DMatch} \small Each row shows the registration result on two RGB-D scans. Underneath each figure, we show the rotation and translation errors of the scans. We compare \texttt{LLT}, and \texttt{SyncM} to \texttt{Ours}. \texttt{GT} is the ground truth.}
\label{tab:qual-results}
\end{table*}

\subsubsection{Adding Modules to Other Methods}
\label{subsec:add-analysis}

\label{subsec:individual_modules}

\begin{figure}[h!]
     \centering
     \begin{subfigure}[b]{0.43\textwidth}
         \centering
         \includegraphics[width=\textwidth]{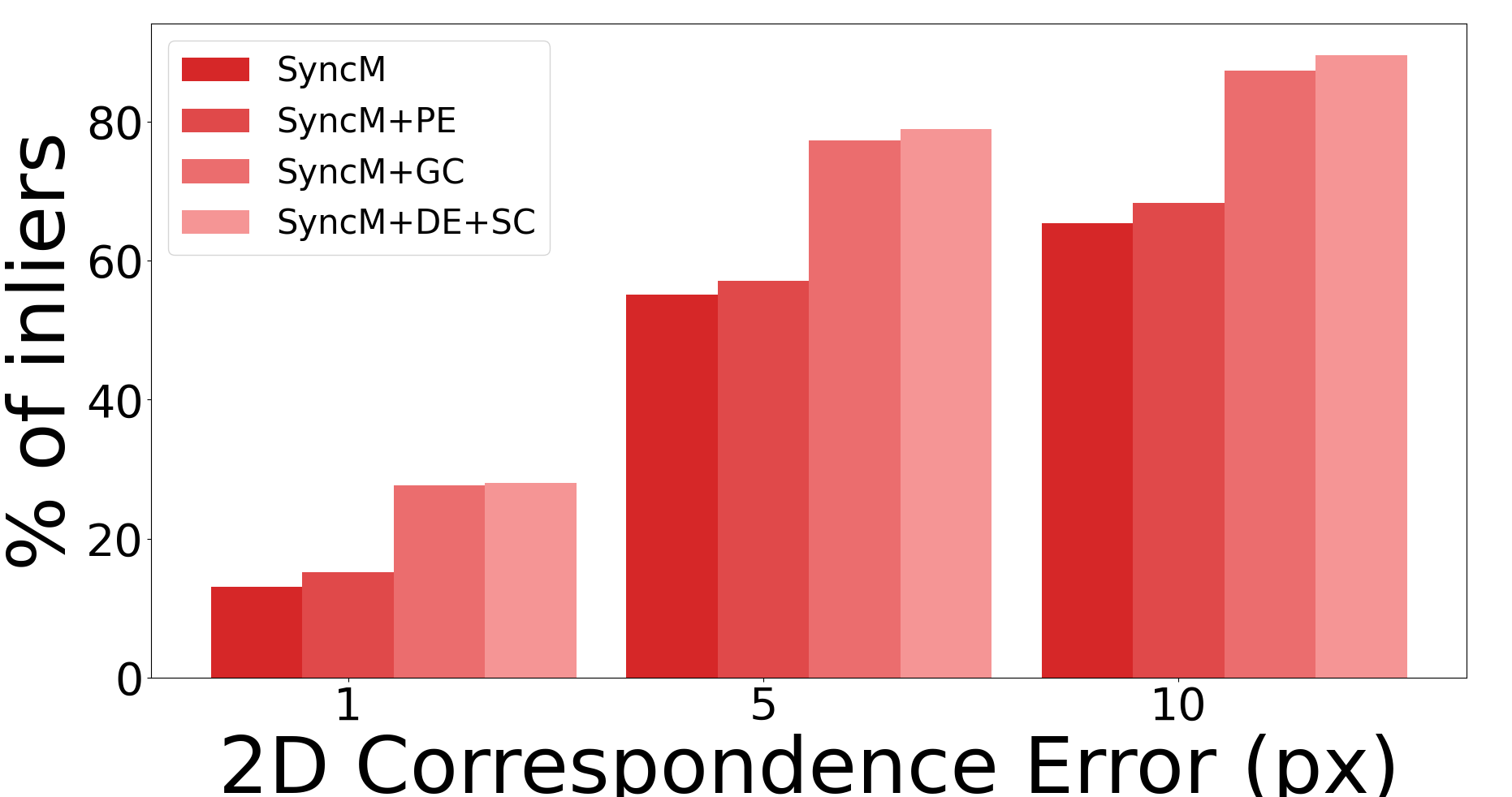}
         \caption{\small 2D Error (in px) }
         \label{fig:corr_2d}
     \end{subfigure}
     \begin{subfigure}[b]{0.43\textwidth}
         \centering
         \includegraphics[width=\textwidth]{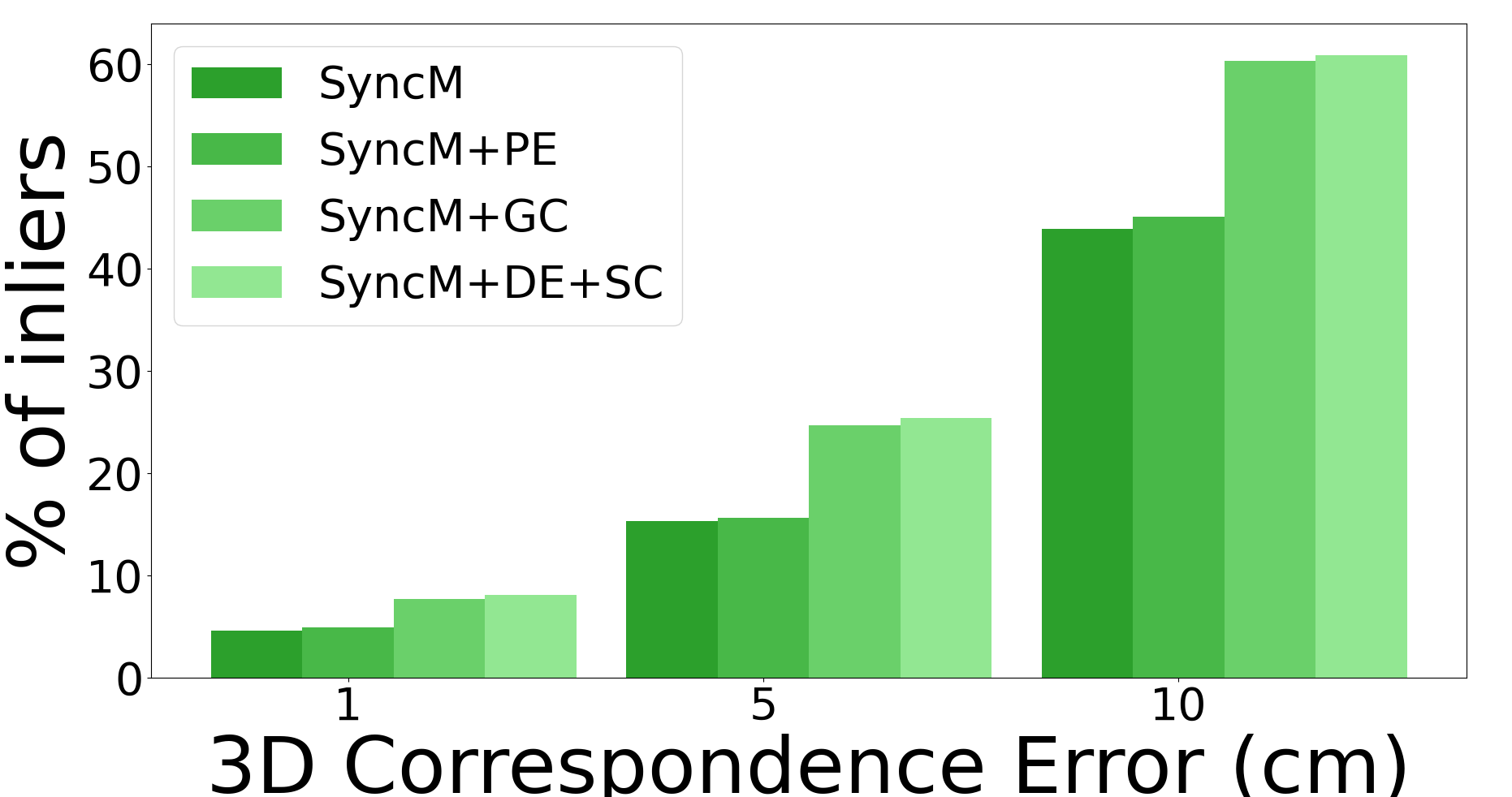}
         \caption{3D Error (in cm)}
         \label{fig:corr_3d}
     \end{subfigure}
\caption{ \textbf{Addition Analysis on Correspondence Error} Adding \texttt{SC+DE} to \texttt{SyncM} has an even more significant impact on inlier \% than even \texttt{GC}. All modules do improve the performance of \texttt{SyncM}.}
\label{fig:syncm_add_corr}
\end{figure}

We also validate the individual proposed modules by adding them to  \texttt{SyncM}. We only do so to \texttt{SyncM.} as \texttt{UR\&R} is quite close to it in methodology and results should be similar for both. While both \texttt{LLT} and \texttt{BYOC} process depth in a significantly different way making the incorporation of our modules in them difficult.

Even modifying \texttt{SyncM} to use our modules is non-trivial. 

We create three modifications of \texttt{SyncM}: \texttt{SyncM + PE} by swapping its pose block with $\texttt{PE}$, \texttt{SyncM + GC} by adding $\texttt{GC}$ to the matching problem and  \texttt{SyncM + DE + SC} where we add \texttt{DE} to the features of SyncM and add \texttt{SC} to the matching problem. Note that the anchor points used in \texttt{SyncM + DE + SC} do not have the noise of our pipeline and are thus not optimized. Thus for \texttt{SyncM + DE + SC}, we only run the inner iteration.

\Cref{tab:rot_ablation} shows the impact of adding the modules SyncM. As can be seen all modules are beneficial to the algorithm. Unsurprisingly \texttt{SyncM + DE + SC} outperforms even \texttt{SyncM + GC}. This is probably because of perfectly localized anchor points. \Cref{fig:syncm_add_corr} mirrors a similar story for correspondence error with \texttt{DE + SC} outperforming \texttt{GC}.

\emph{We show additional ablation studies in the supplementary.}

\begin{table}[h]
\setlength{\tabcolsep}{2pt}
\scriptsize
    \centering
    \begin{tabular}{cccccc|cccc}
       Study& Method & \multicolumn{4}{c}{Angular Error} & \multicolumn{4}{c}{Translation Error}  \\
     &   & \multicolumn{2}{c}{\underline{\text{Accuracy $\uparrow$}}} & \multicolumn{2}{c}{\underline{\text{Error $\downarrow$}}} & \multicolumn{2}{c}{\underline{\text{Accuracy $\uparrow$}}} & \multicolumn{2}{c}{\underline{\text{Error $\downarrow$}}}  \\
 &         & $5^\circ$ & $10^\circ$ & Mean & Med. & $5$ & $10$ & Mean & Med. \\
         \hline
         \\
    \parbox[t]{2mm}{\multirow{5}{*}{\rotatebox[origin=c]{90}{Ablation}}} &\texttt{Ours-DE} & 96.7 & 98.1 & 2.0 & 0.8 & 81.6 & 92.5 & 6.9 & 2.1\\
   & \texttt{Ours-PE} & 96.4 & 97.8 & 2.2 & 0.7 & 83.4 & 91.6 & 6.7 & 2.2\\
   &\texttt{Ours-SC} & 95.3 & 97.7 & 2.4 & 0.7 & 82.1 & 90.7 & 5.1 & 2.3\\
   &\texttt{Ours-GC} & 94.2 & 97.9 & 2.4 & 0.7 & 80.4 & 90.4 & 5.7 & 2.4\\
   & \texttt{Ours} &  \textbf{97.1} & \textbf{98.2} & \textbf{1.9} & \textbf{0.6} & \textbf{85.9} & \textbf{93.6}  & \textbf{3.9} & \textbf{1.9}\\
\\
\hline
\\
     {\multirow{5}{*}{\rotatebox[origin=c]{90}{Addition}}}  
 & \texttt{SyncM} &  93.4 & 97.6 & 2.8 & 0.7  & 76.6 & 89.9 & 7.1 & 2.6 \\
 & \texttt{SyncM+PE} &  92.7 & 97.8 & 2.6 & 0.7  & 79.6 & 90.8 & 5.9 & 2.5 \\
& \texttt{SyncM+DE+SC} &  95.5 & 98.1 & 2.0 & \textbf{0.6}  & 81.3 & 92.1 & 5.4 & 2.1 \\
& \texttt{SyncM+GC} &  95.7 & 97.9 & 2.0 & \textbf{0.6}  & 81.6 & 91.8 & 5.2 & 2.2 \\
\end{tabular}
\caption{ \textbf{Registration Ablation And Addition Analysis}  Best results are bold. Removing \texttt{GC} has the greatest impact on registration error, followed by \texttt{SC}. In the addition analysis, adding \texttt{DE+SC} to \texttt{SyncM} has an even stronger impact than \texttt{GC}. The impact of \texttt{PE} is significant but not dominant.} 
    \label{tab:rot_ablation}
\end{table}

\section{Conclusions}
Leveraging easily identifiable salient portions within RGB-D scenes remains an underexplored resource for geometric reasoning. By using these salient points (anchor points), we constrained the correspondence matching problem, improving correspondence localization. Additionally, we introduced technical enhancements to the registration pipeline, effectively leveraging complementary data sources and enhancing registration accuracy. Consequently, our approach sets a new RGB-D registration state-of-the-art for both ScanNet and 3DMatch benchmarks.

\printbibliography

\addtolength{\textheight}{-12cm}   

\end{document}